\documentclass{article}

\usepackage[preprint]{neurips_2026}

\usepackage[utf8]{inputenc} 
\usepackage[T1]{fontenc}    
\usepackage[hidelinks]{hyperref}       
\usepackage{url}            
\usepackage{booktabs}       
\usepackage{amsfonts}       
\usepackage{nicefrac}       
\usepackage{microtype}      
\usepackage{xcolor}         
\usepackage{amsmath}
\usepackage{amssymb}
\usepackage{mathtools}
\usepackage{amsthm}
\usepackage{comment}
\usepackage[normalem]{ulem}
\usepackage{dsfont}
\usepackage{tcolorbox}
\usepackage{xcolor}
\usepackage{algorithm}
\usepackage{algorithmic}
\usepackage{listings} 
\usepackage{subcaption}
\newif\ifhighlightrevisions
\highlightrevisionstrue
\makeatletter

\newcommand{\old}[1]{\ifhighlightrevisions\expandafter\@firstofone\else\expandafter\@gobble\fi{{\color{red}#1}}}
\makeatother

\lstdefinestyle{jsonstyle}{
    basicstyle=\ttfamily\small,
    breaklines=true,
    frame=none,
    showstringspaces=false,
    keywordstyle=\color{blue},
    stringstyle=\color{red},
    commentstyle=\color{green!50!black},
}

\newtcolorbox{promptbox}{
    colback=gray!10,     
    colframe=gray!50,    
    boxrule=0.5pt,       
    arc=3pt,             
    left=6pt, right=6pt, top=6pt, bottom=6pt,
}

\theoremstyle{plain}
\newtheorem{theorem}{Theorem}[subsection]

\theoremstyle{definition}
\newtheorem{definition}[theorem]{Definition}

\theoremstyle{remark}

\usepackage[textsize=tiny]{todonotes}
\usepackage[acronym]{glossaries}

\newacronym{acr:ai}{AI}{Artificial Intelligence}
\newacronym{acr:llm}{LLM}{Large Language Model}
\newacronym{acr:cpl}{CPL}{Centipawn Loss}
\newacronym{acr:acpl}{ACPL}{Average Centipawn Loss}
\newacronym{acr:ood}{OOD}{out-of-distribution}
\newacronym{acr:nd}{ND}{near-distribution}
\newacronym{acr:wd}{WD}{within-distribution}
\newacronym{acr:td}{TD}{Temporal Difference}
\newacronym{acr:off_tde}{Off-TDE}{Offline Temporal Difference Error}
\newacronym{acr:on_tde}{On-TDE}{Online Temporal Difference Error}
\newacronym{acr:mdp}{MDP}{Markov decision process}
\newacronym{acr:tarl}{TARL}{Target-Aligned Reinforcement Learning}
\newacronym{acr:uci}{UCI}{Universal Chess Interface} 
\makeglossaries
\title{Disentangling generalization and memorization in large language models using chess}

\author{%
  Leonard S.~Pleiss\\
  Technical University of Munich\\
  \texttt{leonard.pleiss@tum.de} \\
  \AND
  Maximilian Schiffer \\
  HEC Paris\\
  \texttt{schiffer@hec.fr} \\
  \And
  Robert K. von Weizsäcker \\
  Technical University of Munich\\
  \texttt{weizsaecker@tum.de} \\
}

\makeatletter
\newcommand\blfootnote[1]{%
  \begingroup
  \renewcommand\thefootnote{}%
  \long\def\@makefntext##1{\noindent##1}%
  \footnote{#1}%
  \addtocounter{footnote}{-1}%
  \endgroup
}

\begin{document}
\newcommand{\idxCurrent}{c}
\newcommand{\idxPrimary}{t}
\newcommand{\idxSecondary}{i}
\newcommand{\idxTertiary}{m}
\newcommand{\idxOracle}{l}
\newcommand{\idxSelected}{j}

\maketitle

\begin{abstract}
Large Language Models (LLMs) exhibit remarkable capabilities, yet it remains unclear to what extent these reflect sophisticated recall or genuine reasoning ability. We introduce chess as a controlled testbed aimed at disentangling these faculties. Leveraging the game’s structure and scalable engine evaluations, we construct a taxonomy of positions varying in density of relevant priors---ranging from common states solvable by memorization to completely novel ones requiring generalization. Crucially, our approach achieves this distinction without requiring explicit knowledge of the models' training data. Applying this taxonomy, we combine a longitudinal analysis of the GPT lineage with a rigorous evaluation of contemporary models, including Claude Opus and Gemini. Our analysis reveals a steep gradient: performance consistently degrades as the density of relevant priors decreases. Notably, for tasks with few relevant priors, base model performance regresses to the random-play baseline\-{, and all models propose illegal moves at sharply increasing rates, indicating that they fail to reliably apply the fixed rules of the game in unfamiliar positions.} While newer models improve, progress slows significantly for tasks with sparse priors. Furthermore, while reasoning-augmented inference improves performance, its relative marginal benefit per token decreases in the absence of relevant priors. These results suggest limitations in systematic generalization, highlighting the need for mechanisms beyond scale to achieve robust performance when deprived of relevant priors.\blfootnote{Data and code are publicly available at \url{https://github.com/leonardpleiss/chess_llm_generalization}.}
\end{abstract}

\section{Introduction}

\glspl{acr:llm} have achieved remarkable progress in recent years, setting new state-of-the-art results across a broad range of benchmarks and applications~\citep{minaee2025largelanguagemodelssurvey}. This rapid advancement has led to widespread adoption in both research and industry~\citep{raza_industrial_2025, llm_survey}. Yet, the origins of these performance gains are still under debate. It remains unclear whether they stem from enhanced retrieval across ever-expanding training corpora, or from improvements in genuine problem-solving and reasoning capability. This distinction is crucial: while both can yield impressive scores on specific benchmarks, only capabilities linked to genuine problem-solving ability enable an agent to systematically reason beyond its prior experience and generalize to novel problems ~\citep{chollet2019_intelligence}. Understanding how much \glspl{acr:llm} rely on memorization, and to which extent they are capable of generalizing beyond their training corpus therefore carries profound implications for their safety, interpretability and further trajectory.

Empirically disentangling memorization and generalization poses a major challenge. It requires evaluating models on tasks that vary in availability of relevant priors: Performance on \gls{acr:wd} tasks, i.e., tasks that were contained in the training data, is generally considered indicative of memorization capabilities, whereas \gls{acr:ood} performance is considered indicative of generalization \citep{chollet2019_intelligence, meszaros2025outofdistributiontestsrevealcompositionality, yang2025trulyassessingfluidintelligence, wu2025understandingllmsfluidintelligence}. 

Yet, because modern \glspl{acr:llm} are trained on massive and largely opaque datasets, quantifying the availability of relevant  priors typically remains out of reach. As a result, existing studies into the origin of \gls{acr:llm} capabilities frequently conflate both concepts, yielding fragmented evidence and sparking an ongoing debate (see Section~\ref{sec:related_work}).

We contribute to this debate by proposing a controlled, empirical framework that leverages chess as a structured environment for studying strategic reasoning in \glspl{acr:llm}. As a domain combining clear rules, deep combinatorial structure, and computationally verifiable  performance metrics via chess engines, chess provides a rigorous testbed for disentangling memorization and generalization: All chess games originate from a single initial state, and many early game positions are extensively studied. Yet, the number of reachable configurations grows combinatorially---approximately $10^{43}$ distinct legal board states exist~\citep{shannon1950}---meaning that novel positions frequently arise after only a few moves. Therefore, chess engages both memorization and generalization: play on known, theoretical positions benefits from the ability to recall known positions, whereas succeeding in novel positions demands generalization beyond memorized game states and patterns (see Appendix~\ref{sec:case_study}). 

In this paper, we conduct a systematic evaluation of \gls{acr:llm} performance in the structured domain of chess across tasks that require differing degrees of memorization and generalization. We do not seek to validate or dismiss \gls{acr:llm} capabilities, but to provide rigorous, domain-grounded evidence that can inform the ongoing discourse on the origin of \gls{acr:llm} capabilities and their developmental trajectory.

Our results reveal a consistent performance gradient across all tested models: performance declines as the density of relevant priors decreases. This observation corroborates recent literature~\citep{qi2024quantifyinggeneralizationcomplexitylarge}. We extend these foundations in four ways. First, without reasoning augmentation, model performance collapses to random levels when relevant priors are sparse. Second, the rate of illegal moves rises sharply as relevant priors become sparser, even though the rules that determine legality are identical in every position; we regard this as our most robust evidence of a generalization failure. Third, while newer model generations improve steadily, the gains from one generation to the next diminish, most strongly for tasks with few relevant priors. Fourth, reasoning-augmented inference provides measurable benefits, but its relative gain per reasoning token decreases when relevant priors are sparse. Collectively, these findings highlight persistent weaknesses in state-of-the-art \glspl{acr:llm} when generalizing beyond the training data manifold.

\section{Related work}\label{sec:related_work}

With this investigation, we aim to contribute to an active debate on the extent to which \glspl{acr:llm} exhibit memorization versus emergent reasoning and generalization. An increasing number of studies explore the degree to which current \glspl{acr:llm} can effectively solve tasks beyond their training data distribution. However, the evidence remains inconclusive. 

Some results point to genuine generalization---or even emergent capabilities~\citep{wei2022emergentabilitieslargelanguage, Ganguli_2022, srivastava2023imitationgamequantifyingextrapolating, brown2020languagemodelsfewshotlearners, si2024llmsgeneratenovelresearch, bubeck2023sparksartificialgeneralintelligence, greatrix2024largelanguagemodelscreate, georgiev2025mathematicalexplorationdiscoveryscale}.  
For instance, \citet{si2024llmsgeneratenovelresearch} found that \gls{acr:llm}-generated research ideas were rated as more novel than those of human experts, while \citet{greatrix2024largelanguagemodelscreate} observed that \glspl{acr:llm} succeed in spatial reasoning tasks with negligible data overlap.
Conversely, other work highlights persistent limitations in reasoning or generalization~\citep{chollet2019_intelligence, chollet_2025_arcagi2, yang2025trulyassessingfluidintelligence, duan2024gtbenchuncoveringstrategicreasoning}.  
At its release, frontier models scored below 1\% on the Abstraction and Reasoning Corpus ARC-AGI-3~\citep{foundation2026arcagi3newchallengefrontier}. Moreover, \citet{yang2025trulyassessingfluidintelligence} demonstrate that while \glspl{acr:llm} excel in low-level cognitive tasks, they struggle in higher-order reasoning settings.  
Such findings suggest that, despite impressive in-context learning and retrieval capabilities, \glspl{acr:llm} may still lack the kind of flexible, compositional reasoning required to meaningfully generalize beyond the training data manifold.

One central reason for this ongoing debate and diverging findings is that existing benchmarks often lack explicit high-quality information on task contamination \citep{li2023taskcontaminationlanguagemodels}, which makes it difficult to disentangle memorization and generalization. Without knowledge of the training corpus, attributing performance to generalization remains inherently speculative. The lack of consensus within this debate therefore largely reflects the resulting challenge in reliable measurement and clean experimental design: A rigorous comparison between memorization and generalization in \glspl{acr:llm} requires evaluation across tasks that differ systematically in their proximity to the model’s training data distribution.  
Because the precise content of training corpora remains unknown, most existing studies rely on benchmarks that implicitly conflate recall and reasoning. 

Various papers have recently tackled the challenge to disentangle memorization and generalization in language models, highlighting the subject's enormous practical relevance \citep{wu2025understandingllmsfluidintelligence, yang2025trulyassessingfluidintelligence,
qi2024quantifyinggeneralizationcomplexitylarge,
xie2025memorizationlargelanguagemodels, wang2025generalizationvsmemorizationtracing, meszaros2025outofdistributiontestsrevealcompositionality}. Most approaches rely on small-scale transformer models to circumvent the issue of training corpus opacity. However, as scale is known to fundamentally alter the dynamics within language models \citep{wei2022emergentabilitieslargelanguage}, it is unclear how these findings generalize to large-scale \glspl{acr:llm}.

We posit that the game of chess offers a unique opportunity to disentangle memorization and generalization in state-of-the-art \glspl{acr:llm}. Chess has played a foundational role in the history of artificial intelligence~\citep{heath1997_history_chess, schaeffer_kasparov_1997}. Yet, the work at the intersection of language modeling and chess thus far primarily focused on general benchmarking or improving domain-specific performance through finetuning and supervised architectural specialization~\citep{wang2024_finetuning-for-chess, ruoss2024amortizedplanninglargescaletransformers, hwang2025_knowledge-dist, chess-llm-performance, kolasani2025llmchessbenchmarkingreasoning}. Only a single work employs chess to investigate generalization versus memorization in transformer architectures \citep{meszaros2025outofdistributiontestsrevealcompositionality}. It relies on a relatively small-scale model, trained via supervised learning, limiting its insights into how their findings generalize to state-of-the-art \glspl{acr:llm}. It therefore remains open how general-purpose frontier \glspl{acr:llm} perform zero-shot as positions become increasingly unfamiliar (see Appendix~\ref{sec:extended_related_work} for a more detailed literature review).

Contrary to this existing stream of research, we employ the unique characteristics of the chess domain to disentangle memorization and generalization in large-scale state-of-the-art language models. The game's combinatorial structure, its well-defined rules and extensive corpora of games enable us to estimate the density of relevant priors in opaque, proprietary training corpora. This unique characteristic facilitates a principled analysis of how contemporary general-purpose \glspl{acr:llm} operate within and beyond their training distribution, and yields valuable insight into the intellectual nature of these systems.

\section{Methodology}\label{sec:methodology}
To systematically evaluate \gls{acr:llm} capabilities as a function of the availability of relevant priors, we employ chess as a controlled testbed combining combinatorial complexity with scalable engine evaluation. This setup allows us to disentangle memorization and generalization by varying positional familiarity. We subsequently (i)~formalize the link between performance under varying densities of relevant priors, memorization and generalization, (ii)~instantiate this framework in chess, (iii)~generate \gls{acr:wd}, \gls{acr:nd}, and \gls{acr:ood} datasets, (iv)~define performance metrics, and (v)~detail the experimental protocol.

\textbf{Conceptual framework.} We study the origin of domain-specific ability within \glspl{acr:llm} through the lens of \emph{memorization} and \emph{generalization}. Following the conceptual framework proposed by \citet{chollet2019_intelligence}, we align this distinction with machine learning terminology by formalizing these concepts in terms of performance under different data distributions.

\begin{definition}[Memorization and generalization in \glspl{acr:llm}]
Let $p_{\text{train}}(x)$ denote the (unknown) training data distribution of an \gls{acr:llm} and let $p_{\text{test}}(x)$ be a test distribution over inputs $x \in \mathcal{X}$ within a given domain.  
We define:

    \emph{Memorization} as the model's ability to perform accurately under \emph{\gls{acr:wd}} conditions, i.e., when $x \sim p_{\text{train}}(x)$. This reflects the model's ability to solve tasks when relevant priors are available, predominantly through retrieval and recall of previously observed solutions.

    \emph{Generalization} as  the model's ability to perform accurately under \emph{\gls{acr:ood}} conditions, i.e., when $x \sim p_{\text{test}}(x)$ with $p_{\text{test}} \neq p_{\text{train}}$. It reflects the model's ability to solve tasks when relevant priors are sparse or entirely unavailable, indicating systematic and adaptive first-principle reasoning beyond previously memorized solutions to solve novel tasks within a familiar domain.
\end{definition}

To operationalize this distinction, we quantify an \gls{acr:llm}'s performance ratio between \gls{acr:wd} and \gls{acr:ood} conditions as a \emph{ generalization gap}, 

\vspace{-1em}
\begin{equation}
    R_{\text{gen}} = \frac{\mathbb{E}_{x \sim p_{\text{test}}}[L(f(x))]}{\mathbb{E}_{x \sim p_{\text{train}}}[L(f(x))]},
\end{equation}
\vspace{-1em}

where $L(f(x))$ is a loss or performance metric associated with the model’s output $f(x)$ on input $x$. High \gls{acr:wd} accuracy reflects strong memorization. Accordingly, as $R_{\text{gen}} \to 1$, \gls{acr:ood} performance approaches \gls{acr:wd} levels, signifying stronger generalization. Naturally, this estimate is only meaningful if \gls{acr:wd} performance exceeds random levels.

Quantifying this gap requires an isolated measure of \gls{acr:wd} and \gls{acr:ood} performance in state-of-the-art \glspl{acr:llm}, which is notoriously difficult because the underlying data distribution is both vast and ill-defined. In consequence, the density of relevant priors is generally difficult to estimate. However, we believe that the domain of chess offers a unique opportunity to approximate this density.

\textbf{Approximating the density of relevant priors in chess.} The state space of chess is combinatorially vast, with \citet{shannon1950} famously bounding the state-space complexity at roughly $10^{43}$. Within this space, a small subset of configurations---predominantly opening positions---is extensively documented and frequently encountered. Given the extensive available data (literature, forum discussions, database appearances) on these frequently encountered positions, we assume they are likely covered by large-scale pre-training corpora, and denote them as \gls{acr:wd} positions: Models can likely achieve strong performance through pure memorization. Conversely, the overwhelming majority of valid chess positions are not well-studied, indicating that they may not be solved via direct recall. However, the fact that a given position is not well-known does not guarantee that there are few relevant priors: Many configurations maintain significant structural homology with familiar states, allowing models to succeed via interpolating known heuristics, which combines elements of memorization and generalization. We term these \gls{acr:nd} positions. To rigorously assess generalization and compel true first-principles reasoning, the evaluation distribution must employ positions that minimize structural similarity to the training manifold. We denote these as \gls{acr:ood} positions.

\textbf{Position categories.} To evaluate \gls{acr:wd},  \gls{acr:nd} and \gls{acr:ood} performance, we construct three subsets of the chess state space, $\mathcal{X}_{\text{WD}}, 
\mathcal{X}_{\text{ND}}, \mathcal{X}_{\text{OOD}}$, each corresponding to a different estimated density of relevant priors within the \gls{acr:llm}’s training distribution. We provide a detailed empirical basis for the condition-specific design principles and exact construction rules in Appendix~\ref{sec:method_app}. 

\textit{Within-distribution positions ($\mathcal{X}_{\text{WD}}$).}
These are well-known positions that occur frequently in human play and are likely present within the training data. We assign a position $x \in \mathcal{X}$ to $\mathcal{X}_{\text{WD}}$ if it occurs at least $n_{\min}=1{,}000$ times in the Lichess Masters database~\citep{lichess_database}. 
Such positions typically arise within extensively studied openings lines and therefore approximate \gls{acr:wd} conditions, $p_{\text{train}}(x_{\text{WD}}) \approx \max_{x \in \mathcal{X}} p_{\text{train}}(x)$.

\textit{Near-distribution positions ($\mathcal{X}_{\text{ND}}$).}
\gls{acr:nd} positions cover an intermediate regime between \gls{acr:wd} and \gls{acr:ood} states. They represent configurations that are \emph{absent} from the Lichess Masters database yet retain some structural characteristics commonly encountered in \gls{acr:wd} states (e.g., material balance, piece development, pawn structure). We generate these by performing $k{=}10$ random legal moves (20 plies) from the starting position and discarding any position that appears in the database. \gls{acr:nd} positions test associative recall, i.e., the recall of solutions from structurally similar positions, which combines elements of generalization and memorization.

\textit{Out-of-distribution positions ($\mathcal{X}_{\text{OOD}}$).}
\gls{acr:ood} positions represent states with few relevant priors. To minimize the utility of associative recall, we aim to minimize the structural similarity to positions likely encountered during training. We construct \gls{acr:ood} positions by placing both kings and up to ten randomly drawn pieces per side on random vacant squares, subject to position validity constraints (i.e., exactly one king per side, no impossible pawn placements, the side not to move cannot be in check). As intended, this creates valid yet atypical board states that are extremely unlikely or outright impossible to arise during human play, forcing the agent to reason from first principles instead of relying on known heuristics.

\begin{figure}
    \centering
    \includegraphics[width=1\linewidth]{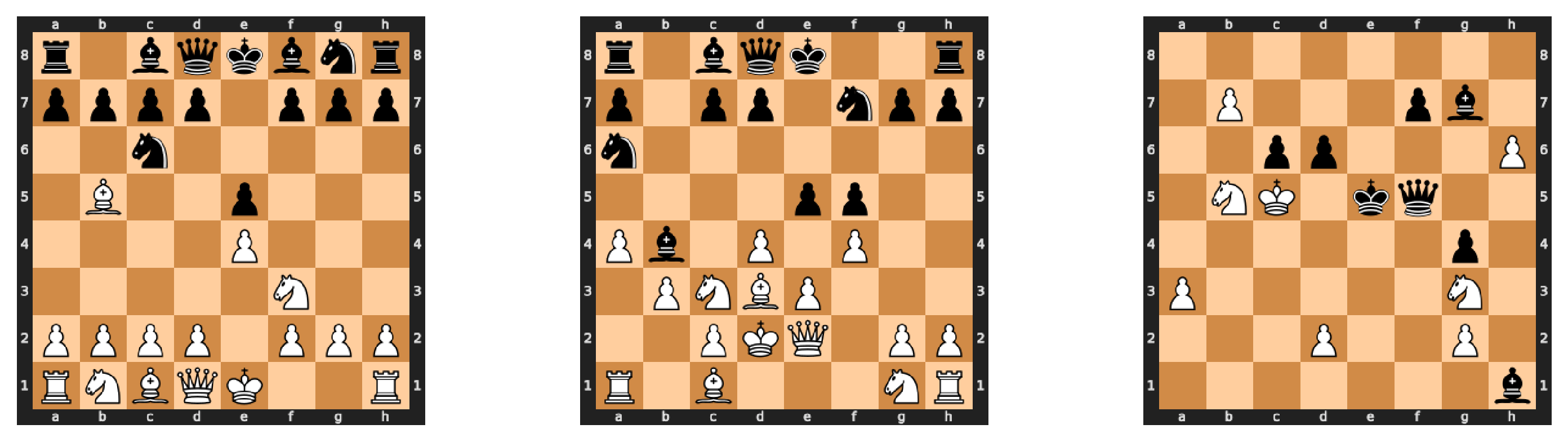}
    \caption{Example positions. Left to right: within-distribution, near-distribution, out-of-distribution. For a detailed discussion of our findings in these positions, we refer to Appendix~\ref{sec:example_positions}.}
    \label{fig:pos}
\end{figure}

\textit{Dataset summary.}
For each condition, we generate N=$500$ positions using the \texttt{python-chess} library~\citep{python-chess},  yielding a balanced dataset of $1{,}500$ board configurations. We refer to Figure~\ref{fig:pos} and Appendix~\ref{sec:example_positions} for example positions. 

\textbf{Move evaluation.} To quantify the quality of model-recommended moves, we employ the \gls{acr:cpl}, a standard measure of decision regret in the chess domain. Unlike other metrics like Win/Loss/Draw Probability, the CPL provides a meaningful signal on move quality, even in imbalanced positions. The \gls{acr:cpl} measures the positional value lost through a move. A centipawn corresponds to one hundredth of a pawn's nominal value, providing a more granular measure of move quality than other established metrics like win probability. Formally, for a position evaluated by a chess engine before and after a candidate move, we define $\ell_{\text{CPL}} = \min ( \max\!\left(0,\, \text{eval}_{\text{b}} - \text{eval}_{\text{a}}\right), 1{,}000 )$,
where $\text{eval}_{\text{b}}$ and $\text{eval}_{\text{a}}$ denote the evaluations from the moving player's perspective before and after executing the selected move.  
Lower values of $\ell_{\text{CPL}}$ indicate stronger play, while higher values reflect increasing positional deterioration. By definition, the best move in each position has $\ell_{\text{CPL}} = 0$. As a broad empirical guideline,
$\ell_{\text{CPL}} < 10$ signifies moves effectively on par with the engine's recommended move,
$10 \leq \ell_{\text{CPL}} < 50$ corresponds to minor inaccuracies,  
$50 \leq \ell_{\text{CPL}} < 100$ to noticeable mistakes,  
and $\ell_{\text{CPL}} \gg 100$ to blunders that substantially reduce winning chances. We acknowledge that CPL is a heuristic metric; however, averaged over a large corpus, it serves as a robust proxy for systematic alignment with superhuman chess engines (see Appendix~\ref{sec:limitations}). In addition to positional quality, we also track the frequency of \emph{illegal moves}, i.e., moves that violate the rules of chess. The proportion of illegal moves serves as a direct indicator of reasoning capabilities---the ability to strictly adhere to formal constraints. According to the rules, playing an illegal move results in instant loss, hence it is scored equally to losing through checkmate, $\ell_{\text{CPL}} = 1{,}000$. We conducted these evaluations using Stockfish~17.1 at search depth~30~\citep{stockfish}, which provides a robust trade-off between computational cost and evaluation reliability.

\textbf{Models.} We analyze the evolution of \gls{acr:wd}, \gls{acr:nd} and \gls{acr:ood} performance across \glspl{acr:llm}. We focus primarily on OpenAI's GPT family to track performance across successive generations of one widely deployed lineage, and evaluate contemporary models from Anthropic and Google to assess whether our findings extend beyond it. We evaluate three GPT generations (Table~\ref{tab:models}): GPT-3.5, GPT-4o, and GPT-5, the most recent GPT generation at the time of data collection. Because GPT-5 allows configurable reasoning effort, we test it at \emph{minimal} and \emph{moderate} effort to assess the impact of reasoning augmentation; we refer to configurations without or with minimal reasoning as \emph{base models}. For the cross-lineage comparison, we evaluate \texttt{gemini-3-flash-preview} and \texttt{claude-opus-4-6} at their lowest available reasoning settings against GPT-5 at minimal effort. As reasoning budgets differ across providers, this constitutes a common low-reasoning baseline that minimizes cross-provider differences in test-time computation, rather than strict parity.

\begin{table}[b!]
  \caption{Overview of model generations, API tags, and release dates}
  \label{tab:models}
  \centering 
  \small     
  \scshape   
  \begin{tabular}{lcc}
    \toprule
      Model Name & API Tag & Release \\
    \midrule
      GPT-3.5 & gpt-3.5-turbo & Mar~2023 \\
      GPT-4o & gpt-4o & May~2024 \\
      GPT-5 & gpt-5 & Aug~2025 \\
      Gemini 3 & gemini-3-flash-preview & Dec~2025 \\
      Claude Opus & claude-opus-4-6 & Feb~2026 \\
    \bottomrule
  \end{tabular}
\end{table}

\textbf{Evaluation protocol.} All GPT responses were collected between September and October~2025 via the OpenAI API; Claude and Gemini responses were obtained in March~2026. Each model received identical prompts requesting a single move recommendation in standard chess notation. As our central quantities are within-model contrasts across conditions, the differing collection windows do not affect them, but they caution against reading small differences between lineages as rankings. Each query--response pair was logged with model metadata and evaluation metrics to ensure full reproducibility.

\textbf{Baselines.}
To contextualize \gls{acr:llm} performance, we introduce two reference points.  
First, we obtain a \emph{random baseline} by uniformly sampling from the set of legal moves in each position.
This represents a lower bound and mirrors play under knowledge of chess rules but without any strategic or tactical capability.
Second, as an upper bound representative of domain-specialized neural network capability, we employ \emph{Leela Chess Zero} (Lc0, ~\citet{leela}), an open-source state-of-the-art neural network chess engine trained via self-play reinforcement learning and derived from the AlphaZero framework~\citep{alphazero}. It employs Monte Carlo Tree Search (MCTS) guided by neural network evaluations, combining pattern recognition with lookahead search. To limit the influence of deep tree search and emphasize the network's evaluative capabilities, we enforce a strict compute budget of two seconds per position. Together, these baselines define the boundaries of a spectrum from rule-constrained random play to domain-optimized neural systems, enabling us to situate \gls{acr:llm} performance across model generations and reasoning configurations within a well-calibrated performance landscape. For details on the configuration, we refer to Appendix~\ref{sec:leela}.

\textbf{Prompting and parsing.} Each chess position $x \in \mathcal{X}$ was encoded using Forsyth-Edwards Notation (FEN) and provided to each model via a standardized prompt requesting the strongest move in Standard Algebraic Notation (SAN) and Universal Chess Interface (UCI) notation, along with a short justification, in a standardized JSON format. The prompt template was held constant across all model configurations to eliminate stylistic bias and isolate systematic differences in capability.  
The template prompt is shown in Appendix~\ref{sec:prompt}. All prompt-response pairs, along with metadata on model identity, token expenditure, and evaluation outcomes, were logged in structured JSON format to enable full reproducibility and downstream analysis.

The recommended moves in SAN notation were checked for syntactic validity. Across all models, syntactically invalid moves occurred in $2.64\%$ of answers ($0.20\%$ for GPT-3.5, $0.07\%$ for GPT-4o, $0.67\%$ for GPT-5 with minimal reasoning, $0.07\%$ for GPT-5 with moderate reasoning, 0.47\% for Gemini, 14.4\% for Claude Opus). Syntactically valid moves that violate the rules of chess in the given position, as well as syntactically invalid moves (including answers from which no move can be extracted), are counted as illegal moves. Moves were parsed with \texttt{parse\_san} from \texttt{python-chess}, which accepts SAN as well as moves given by their start and target squares, as in UCI (e.g., e2e4), also with additional symbols (e.g., g4xg7), but rejects illegal and ambiguous moves. To assess robustness to the output notation, we additionally evaluated the move each model returned in the UCI field of the same response (Appendix~\ref{sec:uci_ablation}). 


\section{Results}

In the following section, we report how model performance evolved across generations, and how it varies across reasoning conditions and model lineages. Unless stated otherwise, all models are evaluated at their lowest available reasoning setting, i.e., minimal reasoning effort for GPT-5, minimal thinking level for Gemini~3, and extended thinking disabled for Claude Opus; GPT-3.5 and GPT-4o offer no reasoning setting. We begin by comparing the average centipawn loss and the frequency of illegal moves across GPT-3.5, GPT-4o, and GPT-5, highlighting consistent but decelerating improvements over time. We then normalize performance metrics to partially control for differences in condition difficulty and compare the observed rates of improvement across generations. Next, we investigate how increased reasoning effort affects GPT-5's chess performance, both in terms of accuracy and computational cost. Lastly, we move beyond the GPT family and compare performance across different state-of-the-art models.

\begin{figure}[t!]

    \centering
    \begin{subfigure}[b]{0.495\textwidth}
        \centering
        \includegraphics[width=\linewidth]{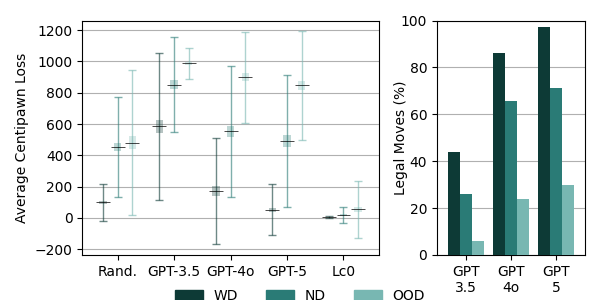}
        \caption{Raw Performance}
        \label{fig:raw_performance}
    \end{subfigure}\hfill 
    \begin{subfigure}[b]{0.495\textwidth}
        \centering
        \includegraphics[width=\linewidth]{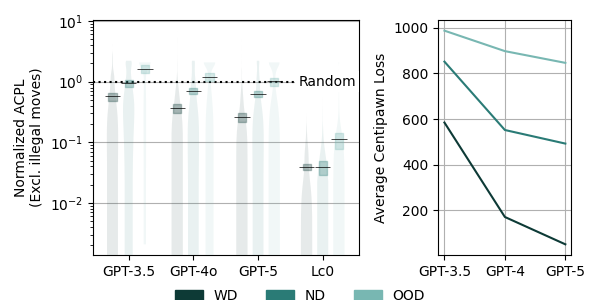}
        \caption{Normalized Performance}
        \label{fig:clean_performance}
    \end{subfigure}
    
    \caption{\textbf{(a)} Left: Average Centipawn Loss with $\text{Matescore} = \text{Illegal Move Score} = 1{,}000$. Right: Proportion of legal moves.  \textbf{(b)} Left: Random-normalized Average Centipawn Loss under exclusion of illegal moves. Right: Observed rates of improvement in Average Centipawn Loss (including illegal moves) across three generations. \textbf{Common parameters:} Horizontal lines indicate means, colored boxes around the mean display the 95\% confidence interval, and vertical lines indicate standard deviations. WD = within-distribution, ND = near-distribution, OOD = out-of-distribution.}
    \label{fig:combined_performance}

\end{figure}

\textbf{Average centipawn loss.} The left panel of Figure~\ref{fig:raw_performance} depicts a consistent increase in average centipawn loss as relevant priors become sparser. Across all models, performance is best for \gls{acr:wd} positions and worst for \gls{acr:ood} ones: the \gls{acr:acpl} in \gls{acr:nd} positions is $4.75\times$ higher than in \gls{acr:wd} positions, and the \gls{acr:acpl} in \gls{acr:ood} positions is $7.79\times$ higher than in \gls{acr:wd} positions. A cross-condition comparison reveals steady but slowing progress: GPT-3.5 achieved an average \gls{acr:acpl} of $807.97$ ($SD=368.58$), GPT-4o of $540.38$ ($SD=461.25$), and GPT-5 of $463.5$ ($SD=463.12$), corresponding to a $33.12\%$ \gls{acr:acpl} reduction from GPT-3.5 to GPT-4o and $14.23\%$ from GPT-4o to GPT-5. The increasing variance across model versions primarily arises from the models' improved ability to avoid exclusively making poor or illegal moves, thereby broadening the distribution of outcomes. Despite this progress, GPT-3.5, 4o and 5 perform worse than random play across all conditions, with GPT-5 in \gls{acr:wd} positions being the sole exception. This indicates that outside of highly familiar contexts, the models struggle to consistently outperform a blind baseline that simply adheres to the rules. This effect is partly due to the high prevalence of illegal moves: The \glspl{acr:llm} frequently incur \glspl{acr:acpl} of $1{,}000$ due to playing illegal moves, whereas the random policy exclusively samples from the set of legal moves, thereby avoiding the penalty.

Lc0, a domain-specialized engine trained via self-play without exposure to human games, achieves a mean \gls{acr:acpl} of $25.10$ ($SD=110.12$), with \glspl{acr:acpl} per condition ranging from $3.91$ ($SD = 5.1$) to $53.73$ ($SD=179.62$), despite limited reliance on explicit search due to the two-second time limit.

\textbf{Illegal moves.}
The right panel of Figure~\ref{fig:raw_performance} reveals that the frequency of illegal moves consistently increases with decrease in relevant priors. Across all model versions, illegal moves are least frequent in \gls{acr:wd} positions and most frequent in \gls{acr:ood} positions. Specifically, on average across models, the proportion of legal moves is $1.46\times$ higher in \gls{acr:wd} positions than in \gls{acr:nd} positions and $4.72\times$ higher than in \gls{acr:ood} positions. Lc0 does not play a single illegal move in any position.

Within the GPT lineage, illegal move rates decrease consistently across successive model versions, suggesting ongoing improvement. GPT-3.5 produced illegal moves in $74.7\%$ of positions, GPT-4o in $41.47\%$, and GPT-5 in $33.8\%$. This corresponds to a decrease in illegal move frequency of $44.46\%$ from GPT-3.5 to GPT-4o and $18.49\%$ from GPT-4o to GPT-5, indicating a deceleration. The high rate of illegal moves in \gls{acr:ood} positions ($\geq 70\%$ for all GPT models) underscores that the difficulty lies not merely in strategic planning, but in the fundamental maintenance of rule consistency when memorized patterns are unavailable. 

\textbf{Normalized average centipawn loss for legal moves.} To isolate pure move quality, we reanalyzed \gls{acr:acpl} under exclusion of illegal moves. We normalized results by dividing each model's legality-filtered \gls{acr:acpl} by the \gls{acr:acpl} of a random legal move policy over this condition. This calibrates each condition against the expected cost of uninformed legal play. In doing so, we obtain normalized \gls{acr:acpl} scores independent of legal move frequency that are comparable across conditions, and partially control for differences in condition difficulty (Appendix~\ref{sec:observational_design}).

The left panel of Figure~\ref{fig:clean_performance} shows that this random-normalized \gls{acr:acpl} replicates the pattern observed previously: for all language models, performance declines monotonically as relevant priors become sparser. Under exclusion of illegal moves, we observe decent performance in positions with relevant priors. For instance, in \gls{acr:wd} positions, GPT-5 achieved a centipawn loss of $26.19$ ($SD=49.53$), outperforming random move selection by a factor of approximately $4$. Notably, even under the exclusion of illegal moves, performance within \gls{acr:ood} positions does not exceed random performance for any of the GPT base models. 

\textbf{Rates of improvement.}
The right panel of Figure~\ref{fig:clean_performance} illustrates that, while models consistently improve, the rate of improvement decreases across generations. Specifically, from GPT-3.5 to GPT-4o, we observed rates of \gls{acr:acpl} improvement of $70.7\%$ for \gls{acr:wd} positions, $35.16\%$ for \gls{acr:nd} positions, and $9.08\%$ for \gls{acr:ood} positions. From GPT-4o to GPT-5, the improvement rates dropped to $69.92\%$, $10.77\%$ and $5.73\%$. Notably, progress slows with decrease in relevant priors in both proportional and absolute numbers. These diminishing returns suggest that scaling current architectures yields progressively smaller generalization gains. While this decrease in return is negligible for \gls{acr:wd} positions ($1.1\%$ reduction), it is substantial in \gls{acr:nd} and \gls{acr:ood} positions ($69.36\%$ and $36.94\%$ reduction).  These diminishing returns suggest that continuous, generation-over-generation refinements---encompassing data curation, post-training alignment, and architectural updates---yield progressively smaller generalization gains.

\begin{figure}[t!]
    \centering
    \begin{subfigure}[b]{0.495\textwidth}
        \centering
        \includegraphics[width=\linewidth]{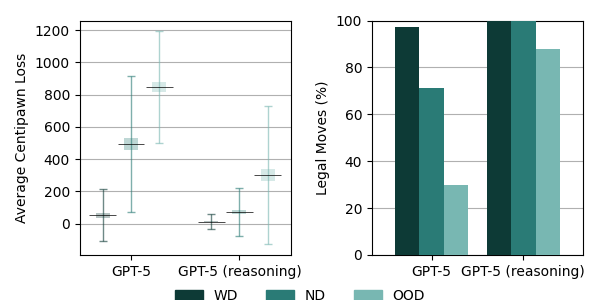}
        \caption{Reasoning impact on performance}
        \label{fig:reasoning_impact}
    \end{subfigure}\hfill
    \begin{subfigure}[b]{0.495\textwidth}
        \centering
        \includegraphics[width=\linewidth]{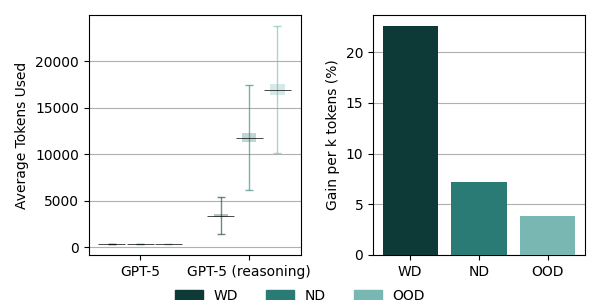}
        \caption{Reasoning cost and gain per token}
        \label{fig:reasoning_effort}
    \end{subfigure}
    
    \caption{\textbf{(a)} Left: Average Centipawn Loss with minimal and moderate reasoning effort. Right: Proportion of legal moves with minimal and moderate reasoning effort. \textbf{(b)} Left: Total tokens used by GPT-5 per position with minimal and moderate reasoning effort. Right: Percentage point improvement per 1,000 tokens by condition. \textbf{Common parameters:} Means as horizontal black lines, 95\% confidence interval as colored boxes, whiskers indicate standard deviations. WD = within-distribution, ND = near-distribution, OOD = out-of-distribution.}
    \label{fig:reasoning_combined}
    
\end{figure}

\textbf{Impact of reasoning.}
We further examined the role of chain-of-thought reasoning. We compared GPT-5 with minimal reasoning effort against GPT-5 with moderate reasoning effort. Figure~\ref{fig:reasoning_impact} illustrates that additional reasoning effort markedly improved performance across all conditions, reducing both average centipawn loss (left panel) and frequency of illegal moves (right panel). While reasoning yields the largest absolute performance gains in the \gls{acr:ood} condition and the smallest in the \gls{acr:wd} condition, absolute differences are inherently confounded by the vastly different baseline performances across conditions. To ensure a calibrated comparison, we therefore focus our subsequent analysis on relative performance gains.  Across conditions, the \gls{acr:acpl} decreased by $72.46\%$. The relative gain was most pronounced in \gls{acr:nd} positions ($85.3\%$ decrease), followed by \gls{acr:wd} ($77.24\%$ decrease) and \gls{acr:ood} ($64.69\%$ decrease) positions. Interestingly, reasoning widens the relative generalization gap: With minimal reasoning, the \gls{acr:acpl} in the \gls{acr:ood} condition was approximately 16 times higher than in the \gls{acr:wd} condition. With moderate reasoning, this gap widens to a factor of approximately 25.

We once again observe that performance follows a clear gradient mirroring the density of relevant priors. Specifically, for \gls{acr:wd} positions, reasoning led to an \gls{acr:acpl} of $11.72$ ($SD=47.44$), reflecting a caliber of play characteristic of advanced human practitioners. Yet, in \gls{acr:ood} positions the model incurs an \gls{acr:acpl} of $298.81$ ($SD=427.61$), which on average corresponds to the loss of a minor piece per move, a deficit that is typically decisive in games between strong players. This observation suggests that even with reasoning enabled, the chain-of-thought process struggles to construct valid strategies in the absence of recognizable pattern anchors.

The left panel of Figure~\ref{fig:reasoning_effort} shows the automatically allocated total number of tokens to increase as relevant priors become sparser: GPT-5 with moderate reasoning effort allocated an average of $3{,}426.41$ ($SD=1{,}967.16$) tokens ($\approx 10.28$ pages, assuming $0.75$ words per token and 250 words per page, \citet{OpenAI_Tokens}) for \gls{acr:wd} positions, $11{,}800.1$ ($SD=5{,}606.32$) tokens ($\approx 35.40$ pages) for \gls{acr:nd} positions, and $16{,}952.54$ ($SD=6{,}831.54$) tokens ($\approx 50.9$ pages) for \gls{acr:ood} ones. This paradoxical increase in computational cost for \gls{acr:ood} tasks---where the model allocates the most resources but achieves the least relative gain---points to an inefficient search process when the solution manifold is unknown.

The right panel of Figure~\ref{fig:reasoning_effort} presents a further analysis revealing that relative performance gains per token diminish as the density of relevant priors decreases: In \gls{acr:wd} positions, $1{,}000$ tokens yield a relative marginal performance gain of $22.54\%$, dropping by $67.93\%$ to $7.23\%$ in \gls{acr:nd} positions, and by $83.07\%$ to $3.82\%$ in \gls{acr:ood} ones. This diminishing relative marginal utility indicates that while reasoning compensates for some deficits, it does not resolve the fundamental challenge of generalizing beyond the training data distribution. Critically, it implies that current chain-of-thought mechanisms act primarily as amplifiers of memorization rather than as independent engines of generalization.

\begin{figure}[t!]
    \centering
    \includegraphics[width=.75\linewidth]{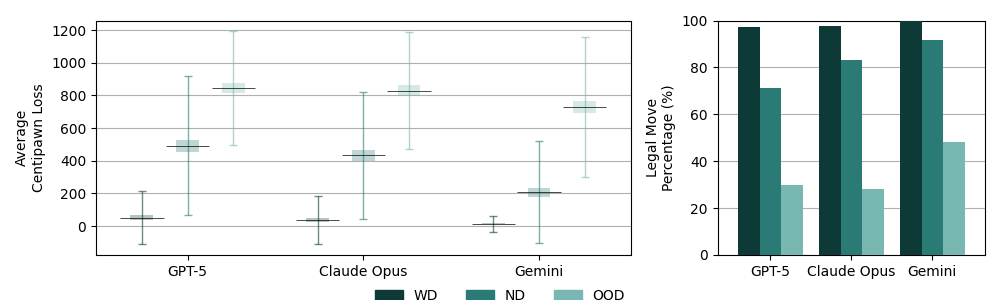}
    
    
    \caption{Performance across model lineages, all at minimal available reasoning effort. Left: Average Centipawn Loss with Matescore = Illegal Move Score = 1{,}000. Right: Proportion of legal moves. Horizontal lines indicate means, boxes around the mean display the 95\% confidence interval, vertical lines indicate standard deviations. WD = within-distribution, ND = near-distribution, OOD = out-of-distribution.}
    \label{fig:other_models}
\end{figure}

\textbf{Performance across state-of-the-art models.} As displayed in Figure~\ref{fig:other_models}, Claude and Gemini replicate the same qualitative performance gradient as the GPT lineage. In the \gls{acr:wd} condition, Gemini achieves an \gls{acr:acpl} of $14.0$ ($SD=49.8$) with an illegal move rate of $0.2\%$, while Claude Opus posts an \gls{acr:acpl} of $34.8$ ($SD=146.8$) with a $2.2\%$ illegal move rate. For the \gls{acr:nd} condition, Gemini records an \gls{acr:acpl} of $207.2$ ($SD=314.0$) alongside an $8.2\%$ illegal move rate, and the \gls{acr:acpl} of Claude Opus shifts to $434.1$ ($SD=390.0$) with $17.0\%$ illegal moves. Finally, in the \gls{acr:ood} condition, Gemini yields an \gls{acr:acpl} of $728.6$ ($SD=427.0$) with $51.6\%$ illegal moves, and Claude Opus reaches an \gls{acr:acpl} of $829.7$ ($SD=359.7$) with an illegal move rate of $72.0\%$. Even these most recent models thus propose illegal moves in more than half of the \gls{acr:ood} positions, compared with at most $2.2\%$ in \gls{acr:wd} positions.

\label{sec:other_models}

\section{Discussion}\label{sec:discussion}

This study examined memorization and generalization, their development across model generations, and reasoning dynamics in \glspl{acr:llm} using the structured domain of chess, and finds a consistent structure in \gls{acr:llm} performance. First, \emph{memorization remains a dominant strength, yet generalization seems brittle, indicating a generalization gap}: When positions resemble those likely encountered during training, models exhibit decent play. However, when relevant priors are sparse, base model performance collapses to random move selection, and the illegal move rate spikes. The steep performance gradient across both illegal moves and move quality suggest that current models have not acquired a robust, generalizable representation of the chess world, but rely on patterns tied to familiar configurations that become brittle once the structural regularities of standard play are removed. We note that the two metrics underlying this finding differ in evidential strength. The rise in illegal moves is the more robust indicator of a generalization failure: legality is governed by rules that are identical in every position, and verifying move legality is a fully specified, bounded procedure. An agent that has internalized these rules should not become less rule-compliant in unfamiliar positions, as the available evidence on human players suggests (Appendix~\ref{sec:human_baseline}). The decline in move quality towards a random baseline is consistent with this trend, but is not conclusively isolated from a potential relationship between novelty and difficulty. Second, \emph{successive model generations show diminishing returns}: although we observe measurable improvements from generation to generation, their magnitude appears to diminish over time. Third, \emph{reasoning helps, but gains diminish when priors are sparse}: Reasoning substantially lowers both illegal move rates and centipawn loss. However, its relative gain per token diminishes as relevant priors become sparser, and the relative generalization gap widens under reasoning. This is consistent with the hypothesis that chain-of-thought mechanisms in their current form function primarily as retrieval optimizers rather than as engines of first-principles derivation. Overall, our results point towards substantial weaknesses in the \gls{acr:ood} performance of \glspl{acr:llm} in chess, and suggest that neither generation-over-generation refinements nor reasoning augmentation, in their current form, reliably extend model competence beyond the training data manifold.

\textbf{Limitations.} Our work comes with multiple limitations. First, our design is observational: Without access to proprietary training corpora, prior density cannot be varied independently of positional structure, so conditions differ jointly in estimated prior density and in properties such as material, board topology, and tactical character (Appendix~\ref{sec:observational_design}). Thus, our findings across model generations and reasoning settings are strictly descriptive, capturing empirical trends rather than establishing universal scaling laws. Second, we cannot guarantee the separation of \gls{acr:wd} and \gls{acr:ood} positions within the training corpus. While we hypothesize that leakage would attenuate the observed gap, we cannot verify this empirically (Appendix~\ref{sec:training_proximity}). Third, despite random-policy normalization and an Lc0 baseline, some confounding between distribution proximity and tactical complexity is unavoidable, narrowing the scope of evaluated generalization primarily toward tactical calculation and rule application rather than broader strategic reasoning (Appendix~\ref{sec:ood_nature}). Fourth, in absence of a human baseline, our move-quality results should be interpreted with caution (Appendix~\ref{sec:human_baseline}). Fifth, our findings replicate across output notations (Appendix~\ref{sec:uci_ablation}), but sensitivity to the input encoding remains untested. Finally, engine-based centipawn loss is imperfect since chess is unsolved, but robust when aggregated over many positions (Appendix~\ref{sec:limitations}).

\textbf{Implications beyond chess.} Chess constitutes a clean, fully observable and analytically rich testbed for studying the interaction between memorization and generalization. The game’s fixed rules, combinatorial structure, and reliable evaluations using superhuman chess engines enable a systematic assessment of generalization without the confounding factors that typically accompany natural language tasks. The cognitive demands required to excel at chess---logically sound reasoning, long-horizon planning, abstract pattern recognition, and exact adherence to formal rules---closely mirror the core competencies necessary to succeed in other complex domains. Consequently, the motifs observed in this study---robust memorization, brittle generalization, and diminishing returns across successive model generations---may not be idiosyncratic to the chess domain. Rather, our findings possibly represent a manifestation of a more general phenomenon in \glspl{acr:llm} that defies measurement in other domains: the systemic difficulty of inferring structural rules to solve novel tasks without close priors.

The diminishing gains observed across successive model generations further suggest that the refinements driving recent progress may not suffice to enable robust generalization. If this pattern extends beyond the present domain, it would imply that model improvements primarily enhance recall and pattern completion within data-rich regions of the training distribution, while leaving structurally novel or low-density regions largely unresolved. Hence, scaling appears to deepen memorized knowledge rather than to broaden fluid adaptability. A similar picture emerges for reasoning. Although chain-of-thought reasoning improves performance, its relative gain per token diminishes as relevant priors become sparser, which challenges the view that reasoning alone can compensate for limited generalization. One interpretation consistent with our data is that extended reasoning chains primarily correct local inconsistencies and surface-level errors rather than inducing new conceptual representations. Under this interpretation, reasoning in current architectures functions mainly as a mechanism for exploiting existing knowledge rather than serving as a reliable pathway toward genuine conceptual synthesis.

Finally, while we anticipate that our findings extend beyond the game of chess, it is crucial to recognize that our empirical findings are limited to a single domain that is characterized by perfect information, and governed by strict combinatorial logic and immense state-space complexity. It remains an open question to what extent the observed limitations transfer to other domains, specifically to highly semantic, probabilistic, or open-ended linguistic ones, where \glspl{acr:llm} frequently demonstrate robust associative flexibility, or to domains receiving extensive targeted post-training.

\textbf{Future work.} Promising extensions include (i) a constrained setting in which models choose from the complete set of legal moves, separating move generation from strategic evaluation; (ii) rare but reachable positions matched on material and legal-move count; (iii) alternative input encodings; (iv) a human baseline on randomized positions; and (v) analyses of reasoning traces.

\clearpage

\begin{ack}
L.S.P. was supported by a Roland Berger doctoral fellowship. Roland Berger had no role in the design, execution, analysis, or reporting of this research. The authors declare no competing interests. We thank Marc Leonard Overbeck (ETH Zurich) for an independent reproducibility check of an earlier version of this paper.
\end{ack}

\bibliographystyle{plainnat}
\bibliography{references}


\newpage
\appendix

\section{Methodological details}\label{sec:method_app}

\subsection{Prompt}\label{sec:prompt}

Positions were passed to the model in Forsyth-Edwards Notation (FEN). The following prompt was utilized to obtain model recommendations with \texttt{\{fen\}} being filled in with the position-specific FEN-string.

\begin{promptbox}
\subsection*{INPUT:}
\noindent Consider the following chess position, denoted as a FEN string: \texttt{\{fen\}}

\subsection*{TASK:}
\noindent Recommend the best move to play in the provided position and state your reasoning in the required format.

\subsection*{OUTPUT REQUIREMENTS:}
\noindent You must return JSON only (no extra text, no code fences, no comments). Return a single JSON object matching this schema:
\begin{lstlisting}[style=jsonstyle]
{
    move_uci: <Your move recommendation in UCI format, e.g. e2e4>,
    move_san: <Your move recommendation in SAN-Format, e.g. e4>,
    reasoning: <Provide a concise rationale why you decided for this move; no more than 3 sentences>
}
\end{lstlisting}

\subsection*{IMPORTANT INFORMATION:}
\begin{itemize}
    \item Your performance is scored on format correctness and recommendation quality.
    \item Not providing a legal chess move in the given position is considered a failed attempt.
    \item Only return JSON. Not precisely following the output requirements is considered a failed attempt.
    \item You are strictly forbidden from accessing external resources, like conducting a websearch or using other tools. This, too, will be considered a failed attempt.
\end{itemize}
\end{promptbox}

\paragraph{Output-token limit.} In contrast to the OpenAI and Gemini APIs, the Anthropic Messages API requires an explicit output-token limit. We set it to 1,024 tokens, which we judged sufficient for the required output (a single move and a rationale of at most three sentences). In practice, 201 of Claude’s 1,500 responses (13.4\%) reached this limit: 200 \gls{acr:ood} positions (40\%), 1 \gls{acr:nd} position and none of the \gls{acr:wd} positions. We score truncated responses that do not provide a legal move like any other response that does not provide a legal move, as truncation results from the model not following the requested output format: the prompt requires a single JSON object with a rationale of at most three sentences and defines any deviation as a failed attempt. All 201 truncated responses began with free-text reasoning before any JSON object, contrary to the instruction to return JSON only. A compliant response is far shorter than the limit: Claude's non-truncated responses that contain a legal move have a median length of 119 output tokens. Excluding truncated responses, Claude’s illegal-move rate is 54.3\% on \gls{acr:ood} positions (n = 300, vs. 72.0\% including them), 16.8\% on \gls{acr:nd} (vs. 17.0\%) and 2.2\% on \gls{acr:wd} (unchanged). The excluded responses are not a random sample, so these figures should not be read as estimates of Claude’s performance without the limit. They do show, however, that truncation alone does not account for the degradation on \gls{acr:ood} positions.

\subsection{Coverage study}\label{sec:case_study}
To empirically assess and challenge our theoretical assumptions about the relationship between game progress and database frequency, we simulated 50 chess games. Each game was generated by iteratively selecting one of the three highest-rated engine moves for both Black and White using Stockfish~\citep{stockfish}. For every resulting position, we queried the Lichess masters database to determine whether it had previously occurred~\citep{lichess_database}. The Lichess Masters database contains more than 2.7 million games played since 1952 of players with FIDE ratings of $2{,}200$ or more. In addition, we tracked the number of pieces remaining on the board, since modern tablebases are known to include all possible positions with seven or fewer pieces~\citep{zakharov_solving_2019}. We therefore consider a position as covered---that is, potentially solvable via memorization--- if it occurs in the Lichess Masters database or contains at most seven pieces. We then computed database coverage rates as a function of move number. 

\begin{figure}[h!]
    \centering
    \includegraphics[width=1\linewidth]{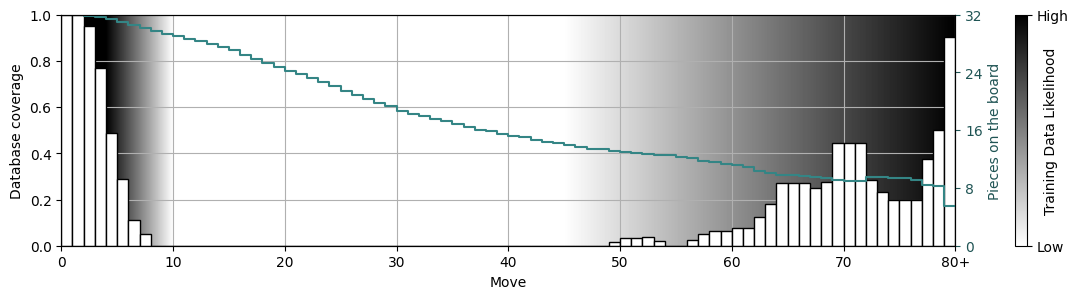}
    \caption{Lichess Masters database coverage by move number.}
    \label{fig:coverage}
\end{figure}

Our results, displayed in Figure~\ref{fig:coverage}, show that coverage is initially high, drops rapidly as the game progresses, and then rises again in later stages---beginning around move 49---as the number of pieces begins falling below eight in some games. We thus conclude that the training data coverage likelihood is high in very early and very late game stages, gradually decreases towards the mid-game and reaches its minimum in the late mid-game, around move thirty, when there are approximately 20 pieces on the board.

\paragraph{Derivation of condition design principles.}
Beyond validating our theoretical assumptions, the coverage study also aimed to inform the derivation of design principles for constructing positions with varying levels of familiarity. Our findings were threefold:

\begin{enumerate}
    \item \textbf{\gls{acr:wd} positions.} 
    High coverage rates can be found in early and certain late game positions. Yet, coverage in late-game positions solely relies on tablebases, where each position occurs once. As such, even if tablebases are part of the training corpora, late-game positions were likely seen few times. In contrast, early-game positions occur thousands of times within the Lichess database, and likely even in other sources such as opening books. Thus, early-game positions exhibit the highest likelihood of frequent and consistent coverage. We therefore obtained \gls{acr:wd} samples by performing a structured breadth-first search (BFS) starting from the initial chess position. We retained only positions that appeared in at least $1{,}000$\footnote{We selected a threshold of $1{,}000$ based on an empirical sensitivity analysis. Lower thresholds (e.g., $100$) introduced significant noise from under-studied positions, while more restrictive thresholds (e.g., $10{,}000$) limited the sample size.} games, thereby maximizing both early occurrence and empirical frequency.

    \item \textbf{\gls{acr:nd} positions.} 
    The empirical finding of novel games beyond move nine in our case study suggested that generating positions by playing ten random moves per side from the starting position yields positions that are not memorized explicitly but remain structurally related to early-game configurations. Importantly, in some modern main lines, moves are known more than thirty moves deep. To ensure that the \gls{acr:nd} position is not part of one of those well-known main lines, we excluded positions which were found within the Lichess database. \gls{acr:nd} positions preserve key strategic features---such as central control, typical pawn structures, and early piece activation---allowing for the application of known heuristics.

    \item \textbf{\gls{acr:ood} positions.} 
    To minimize structural similarity to \gls{acr:wd} configurations, we generated positions via random piece placement rather than random play. Based on our analysis, positions with no database coverage contained on average $20.25$ pieces. We therefore drew ten pieces per side when constructing \gls{acr:ood} positions.
\end{enumerate}

\subsection{Within-distribution dataset generation}

To construct the within-distribution dataset, we systematically extract 500 established opening configurations from the Lichess Masters database via a bounded breadth-first search (BFS) rooted at the standard initial chess position. State representations are tracked using Forsyth-Edwards Notation (FEN). For each dequeued state, we compute its total historical frequency. If a state appears at least $1{,}000$ times, it is appended to the dataset, and its legal continuations are enqueued for further evaluation. Crucially, if a state fails to meet this threshold, it is discarded and its branch is strictly pruned, preventing the exploration of low-frequency variations. The search terminates immediately upon reaching 500 positions, ensuring the final dataset comprises logically contiguous, highly familiar theoretical lines.

\subsection{Out-of-distribution dataset generation}\label{sec:ood_generation}

To generate the \gls{acr:ood} ($\mathcal{X}_{\text{OOD}}$) dataset, we employ a randomized piece placement algorithm utilizing the \texttt{python-chess} library. The objective is to construct board states that are syntactically valid under the formal rules of chess but structurally devoid of typical strategic patterns, making their presence in the models' training corpora highly unlikely.

The generation process iteratively attempts to build positions satisfying several strict constraints. First, regarding king placement, the White and Black kings are placed on two distinct, uniformly sampled squares. Following this, we address piece distribution by drawing 10 additional pieces per side and placing each on a uniformly sampled vacant square; draws that would exceed the maximum count for a piece type are skipped, so each side receives up to 10 additional pieces. The piece types are sampled with weights reflecting standard starting frequencies while discounting pawn probability to account for frequent early-game pawn trades and prevent closed structures, ensuring piece mobility (Pawns: 4, Knights: 2, Bishops: 2, Rooks: 2, Queens: 1). We strictly enforce maximum piece counts corresponding to a standard chess set, ensuring there are no more than 8 pawns or 1 queen per side.

To guarantee positional validity, the board state is verified using \texttt{python-chess} internal validation routines, ensuring fundamental legality such as confirming the side not to move is not currently in check. We immediately discard any position that is already in a terminal state, such as checkmate or stalemate. Finally, for database filtration, the generated FEN string is queried against the Lichess database. The position is solely accepted into the final dataset if it has exactly zero historical occurrences.

\subsection{Leela Chess Zero Configuration}\label{sec:leela}

We utilized a locally deployed instance of the Lc0 network \texttt{t3-\allowbreak 512x15x16h-\allowbreak distill-\allowbreak swa-\allowbreak 2767500}. It represents a medium-sized configuration recommended for both GPU and CPU settings. It employs a hidden dimension of 512 and 15 layers, yielding a GPU memory footprint of approximately 1.8 GB and a compact file size of around 150-155 MB. As a distilled model with stochastic weight averaging (SWA), it captures much of the strength of larger networks while remaining more accessible for resource-constrained environments. This positioning makes it a practical choice for experiments requiring strong playing strength without the computational overhead of very large models \citep{leela}.

\section{Example positions}\label{sec:example_positions}

\begin{figure}[h!]
    \centering
    \includegraphics[width=1\linewidth]{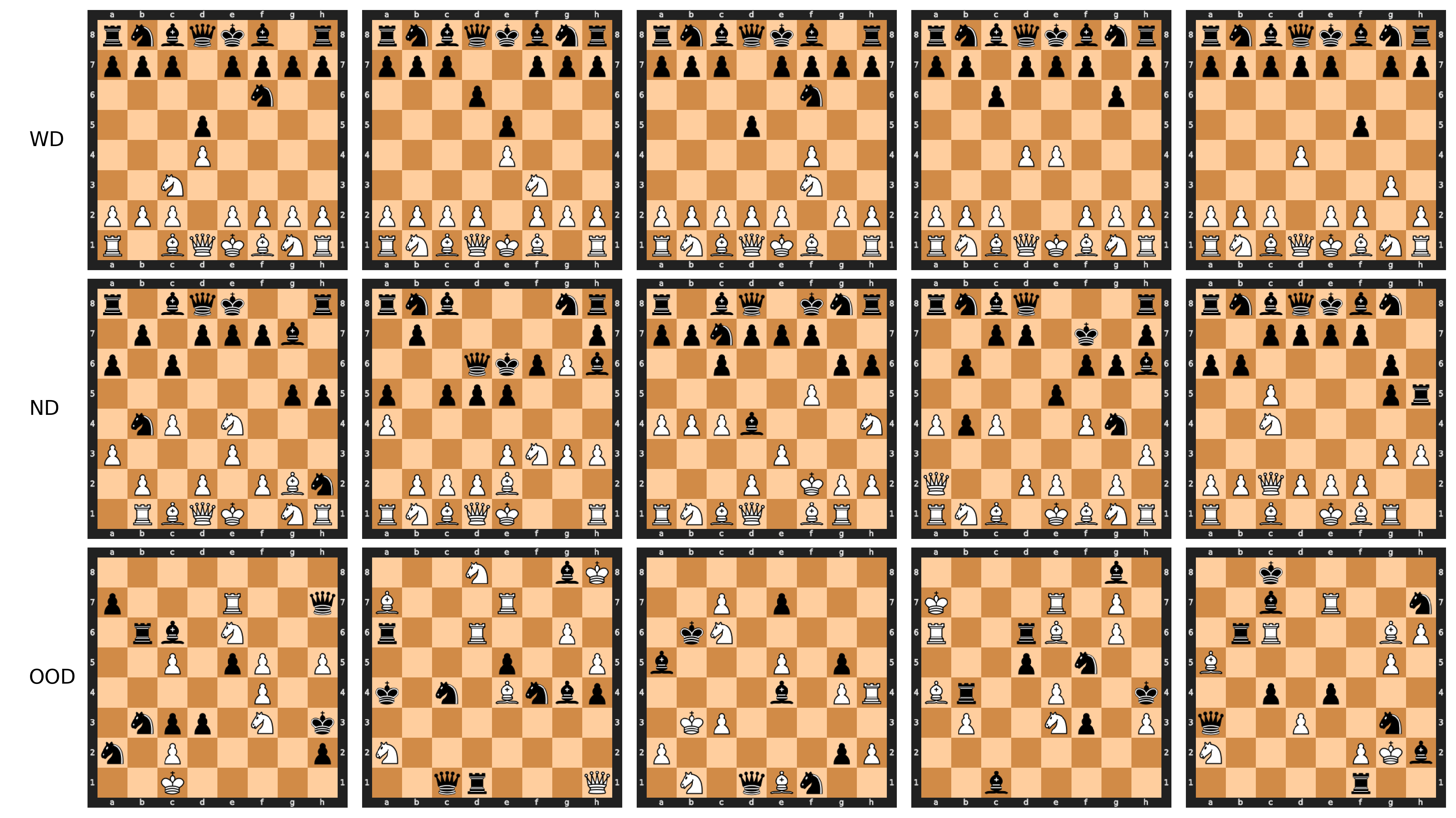}
    \caption{Example positions by condition. WD = within-distribution, ND = near-distribution, OOD = out-of-distribution.}
    \label{fig:example_positions}
\end{figure}

We subsequently discuss the example positions displayed in Figure~\ref{fig:pos} to contextualize our quantitative results.

\paragraph{WD} This is the Ruy López Opening, with black to play. Stockfish recommends ...Nf6, indicating a slight advantage for white (+0.28). GPT-3.5 tries to play ...Qc8, which is illegal. Lc0, GPT-4o and GPT-5 (moderate reasoning) all recommend ...a6, a well-known theoretical continuation known as the Morphy Defense, incurring a negligible CPL of 13. GPT-5 with minimal reasoning recommends Nd4, also following a known theoretical line called the Bird's Defense. It is not considered optimal by modern standards, incurring a CPL of 43. The random agent selects ...Nf6, matching Stockfish's choice.

\paragraph{ND} White is to play. While there are many decent moves in the position, Stockfish recommends Bb2, developing the bishop. After Bb2, white is better (+3.38). Lc0 recommends Bxa6, taking the knight, incurring a CPL of 0. GPT-3.5 recommends to play e4, incurring a CPL of 713, effectively losing the game: After ...exd4, white loses a piece. GPT-4o recommends fxe5, incurring a CPL of 31. GPT-5 (minimal reasoning) tries to play cxd5, which is illegal. GPT-5 (moderate reasoning), like Lc0, recommends Bxa6, also incurring a CPL of 0. The random recommendation is Ba3, incurring a CPL of 88.

\paragraph{OOD} White is to play. Stockfish recommends Kb6 to get out of check. After Kb6, white is better (+3.24), as black cannot stop white from promoting the b7 pawn. Lc0 finds this move, incurring a CPL of 0. GPT-3.5 recommends illegally promoting the pawn via b8=Q, ignoring that white is currently in check. GPT-4o recommends b7g7+, which is not a legal move. GPT-5 (minimal reasoning) recommends playing Nxd4+, also ignoring that white is in check, incurring an illegal move. GPT-5 (moderate reasoning) recommends Nxd6, recognizing the check. While legal, it effectively loses the game to ...Qf2+, as white will no longer be able to promote its pawn, incurring a CPL of 992. The random recommendation is also Nxd6, incurring a CPL of 992.

We provide further example positions in Figure~\ref{fig:example_positions}. A study of the generated positional categories reveals that many positions, especially within the \gls{acr:nd} and \gls{acr:ood} setting, are unlikely or impossible to occur in games between strong players. This systematic difference to practical positions is intentional. We explicitly aim to break known patterns, especially in \gls{acr:ood} positions. In doing so, we aim to minimize the utility of basic pattern recognition and associative recall. Thereby, we decrease the likelihood of obtaining strong solutions using memorized game knowledge like opening theory,  databases, or standard mid- or end-game strategy, aiming to isolate genuine reasoning capability.

\section{Framework validity and scope}\label{sec:limitations}

Our experimental framework rests on several foundational characteristics: the reference world defined by chess engines, the construction of position types, the approximation of training-distribution proximity, the choice of evaluation metrics, the input encoding, the observational nature of the design, and the absence of a human baseline. Each introduces specific constraints and abstractions. We explicitly delineate the scope of our claims below to address potential validity concerns regarding the use of chess as a proxy for general reasoning.

\subsection{The reference world: abstracted engine chess}
It is critical to note that the reference world for this study is not the sport of chess as played by humans, but rather a reduced, abstracted world defined by chess engines. Stockfish and similar engines (including the neural-network-based Lc0) are not intelligent in the human sense. They are highly optimized, heuristic search algorithms rooted in massive calculation \citep{stockfish, leela}. They do not understand chess strategy as a narrative; they solve it as a logic puzzle. Therefore, when we measure \gls{acr:llm} performance against an engine, we are measuring the model's ability to align with a computational optimum, not its ability to exhibit human-like intuition or creativity.

While engines are not ground truth solvers of chess (which remains unsolved), they exceeded human capability decades ago \citep{schaeffer_kasparov_1997}. For the purpose of evaluating \gls{acr:llm} errors---which often involve fundamental misunderstandings of board state or rules---the engine serves as a sufficiently high-resolution ruler. A high \gls{acr:cpl} objectively represents a failure to optimize within the game's logical constraints, regardless of the human quality of the move. Consequently, while engine evaluations may not capture the full semantic nuance of human strategy, they provide a necessary and rigorous baseline for assessing the tactical soundness of the models' reasoning. Thus, within the scope of this study, we accept engine evaluations as a robust heuristic for distinguishing between valid reasoning and retrieval failure, without claiming they represent a ground truth.

\subsection{Nature of OOD positions}\label{sec:ood_nature}
Our \gls{acr:ood} positions are generated via random placement to maximize distance from the training manifold. This generation process introduces specific structural biases that warrant clarification: proficient human play relies on a synergy of \emph{strategy} (often termed \textit{positional play} in domain literature, focusing on long-term planning and static evaluation) and \emph{tactics} (immediate calculation of concrete variations). Due to their randomized nature, our \gls{acr:ood} positions often assess tactical ability rather than strategical prowess. They frequently lack standard pawn skeletons, or King safety norms typical of human games.

While unnatural relative to human play, this tactical focus is a direct consequence of isolating generalization rather than an arbitrary design choice. In chess, familiar patterns are carried predominantly by pawn structures, which form the skeleton of a position and anchor most strategic knowledge. Strategic positions therefore almost always structurally resemble known games, making them highly susceptible to memorization. Breaking familiar patterns requires departing from typical pawn structures---in our generator, by down-weighting pawns (Appendix~\ref{sec:ood_generation})---which in turn renders the resulting positions predominantly tactical. By requiring the model to solve positions that break standard heuristic patterns, we test its ability to apply the rules of the game and derive a solution from the concrete position. In a formal system like chess, when pattern matching is precluded, calculation becomes the primary proxy for fluid reasoning. Accordingly, our claims about generalization primarily concern tactical calculation and rule application rather than long-term strategic planning, and a performance collapse on these positions indicates a limitation in exactly this capacity.

This limitation stands in stark contrast to human capabilities. While human players also rely heavily on structural chunking and heuristic intuition---which are largely inapplicable in \gls{acr:ood} positions---they retain the ability to consistently identify legal moves and calculate local tactics in \emph{any} position. The \glspl{acr:llm}' failure in \gls{acr:ood} settings, highlighted not just by a very high ACPL but by a catastrophic spike in illegal moves, demonstrates an inability to fall back on fundamental rule adherence. This underscores a critical deficit in the foundational components of formal fluid reasoning.

\subsection{Approximation of Training Data Proximity}\label{sec:training_proximity}
The taxonomy employed (\gls{acr:wd}, \gls{acr:nd}, \gls{acr:ood}) relies on an approximation of the unobserved training distribution. We utilized the Lichess Masters database to approximate the training distribution. We acknowledge that unlike curated professional databases (e.g., ChessBase), public internet databases often contain theoretically irrelevant positions. To mitigate this noise, we enforced a strict threshold of $n_{min}=1{,}000$ occurrences for the \gls{acr:wd} set. While this specific cutoff is heuristic, it serves as a robust filter against idiosyncratic games, ensuring that WD positions are not merely present in the database, but are statistically established standard positions likely to be over-represented in the \gls{acr:llm}'s massive training corpora.

We cannot formally guarantee that all positions categorized as \gls{acr:wd} were present in the \gls{acr:llm} training corpus, nor strictly preclude the appearance of \gls{acr:ood} positions. However, given the combinatorial explosion of the state space, the probability of a specific, randomly generated position with high piece count appearing in the training data is negligible. Even if the categorization is imperfect, this uncertainty biases our results conservatively: Misclassification would act as a form of label noise between conditions. \gls{acr:ood} positions that were in fact seen during training would, under the plausible assumption that exposure does not impair performance, raise \gls{acr:ood} performance toward \gls{acr:wd} levels. Conversely, \gls{acr:wd} positions that were in fact absent would lower \gls{acr:wd} performance toward \gls{acr:ood} levels. Both effects attenuate the measured difference between conditions. We therefore hypothesize that the reported results likely represent an upper bound on the model's true fluid reasoning capability, which may be even more limited than observed. We emphasize that this is a reasoned hypothesis rather than an established fact: it cannot be verified empirically without access to proprietary training corpora, and it relies on the assumption that exposure to a position does not systematically harm performance.

\subsection{Choice of evaluation metrics}
We employ \gls{acr:cpl}---a measure of decision regret relative to an engine's optimal move---as our primary metric. Critically, we argue that outcome-based metrics are insufficient for measuring performance in this context, as they fail to distinguish between a reasoning process that is logically sound but imperfect, and one that is effectively random. However, as engines do not exhibit perfect play, we explicitly acknowledge that \gls{acr:cpl} is not a flawless descriptor of overall playing strength, particularly in strategic contexts where sub-optimal moves may result in long-term positional disadvantages that may exceed the evaluation depth of contemporary engines. Our choice is nonetheless motivated by the fact that alternative methods like win/draw/loss probability do not sufficiently differentiate move quality in positions that are objectively decided. This saturation masks qualitative differences in play. Conversely, \gls{acr:cpl} provides a continuous measure of gradual error magnitude. It distinguishes between a sub-optimal move that retains the current balance (e.g., \gls{acr:cpl} 15) and a blunder (e.g., \gls{acr:cpl} 300), even in imbalanced positions. This granularity is essential for quantifying the subtle gradients of reasoning capability across model generations that would otherwise remain invisible under binary evaluation criteria.

We recognize that CPL is a non-linear metric and its magnitude is sensitive to the absolute evaluation of a given position. Because our \gls{acr:ood} positions are generated randomly, they may feature material imbalances from the outset, which can inflate or deflate CPL penalties for sub-optimal moves. However, given the sample size within each condition, we expect these localized positional fluctuations to average out across the dataset, yielding a representative aggregate measure. Nevertheless, to further mitigate the risk of over-interpreting CPL variance, we consider the \gls{acr:acpl} in conjunction with the rate of illegal moves. The concurrent, severe degradation of both metrics in \gls{acr:ood} contexts confirms a fundamental breakdown in positional reasoning, independent of CPL non-linearity.

\subsection{The interplay of tokenization and generalization}

The observed performance degradation in \gls{acr:ood} scenarios may intersect with the input-encoding mechanisms of current \gls{acr:llm} architectures, specifically Byte-Pair Encoding (BPE) tokenizers. Board states in our study were provided via FEN strings. Because modern tokenizers optimize for subword frequencies, standard \gls{acr:wd} FENs may be compressed into familiar, semantically rich tokens due to their high prevalence in the pre-training data. Conversely, the highly atypical FENs generated for our \gls{acr:ood} dataset could force the tokenizer to fracture the input into unlearned, character-level fragments.

While this modality mismatch could be viewed as a perceptual bottleneck rather than a purely strategic failure, it fundamentally reinforces our core thesis regarding the limits of generalization in \glspl{acr:llm}. The reliance on familiar subword tokenization to parse a strictly logical, rules-based environment is itself a manifestation of over-reliance on crystallized patterns. A system possessing robust fluid ability should be capable of falling back on strict syntactic and character-level deduction when familiar heuristic chunks are unavailable. The fact that the base models instead suffer a catastrophic breakdown---evidenced by the spike in illegal moves---demonstrates that their reasoning capabilities are inextricably tethered to the recognition of memorized input distributions. Future work could isolate the exact magnitude of this tokenization effect by evaluating models using direct spatial encodings, though our findings strongly suggest that escaping this crystallized dependency will require architectural mechanisms beyond current autoregressive scaling.

\subsection{Observational design and the limits of causal attribution}\label{sec:observational_design}

We emphasize that our study is observational and does not constitute a randomized controlled trial: prior density is not manipulated independently of the structural properties of the board. Assignment is instead determined by the position-generating process, with the consequence that our conditions differ jointly in estimated prior density, but also in material configuration, pawn structure, board topology, game phase, branching factor, and tactical character. We stress that this is not a control we omitted, but one that the domain itself precludes. Our coverage study (Appendix A.2) demonstrates that structural properties and database coverage are themselves correlated. We argue that constructing positions that are simultaneously low in prior density and structurally indistinguishable from high-prior-density positions is not possible, because \emph{structural atypicality is constitutive of what it means for a chess position to fall outside the human game corpus}. A genuine randomized manipulation would require intervening on the pre-training corpus itself---precisely the object that is inaccessible for proprietary frontier models, and the inaccessibility that motivates our proxy-based design in the first place. Controlling for structural confounds is only feasible employing small-scale, self-trained architectures, at the cost of forfeiting any analysis of frontier-scale behavior. We scope our claims to reflect this limitation: we document a robust association between estimated prior density and performance, but we do not claim to identify a causal effect of prior density net of all co-varying positional properties.

Three features of our design nonetheless constrain the space of alternative explanations. First, the \gls{acr:nd} condition is absent from the database yet structurally continuous with early-game play, and already exhibits substantial, graded degradation relative to \gls{acr:wd}, which is difficult to reconcile with an account in which only extreme structural unnaturalness drives the effect. Second, the rules that determine legality are identical in every position. Legality determination is a bounded procedure whose difficulty does not increase in our \gls{acr:ood} positions. Third, the degradation in legality persists across output notations (Appendix~\ref{sec:uci_ablation}) and under a roughly fivefold increase in reasoning tokens. Jointly, these observations render a purely structural account of our findings difficult to sustain, even though they do not constitute causal identification in the strict experimental sense. In particular, we interpret the legality results as evidence that the models fail to robustly apply the rules of chess in unfamiliar positions, rather than as proof that memorization and generalization have been fully separated.

\subsection{Legality, move quality, and the absence of a human baseline}\label{sec:human_baseline}

A natural objection to our interpretation is that humans, too, may perform worse on unfamiliar positions. Human chess expertise relies heavily on chunking, i.e., the recognition of familiar configurations of pieces~\citep{chase1973perception}, and such chunks are largely unavailable in our \gls{acr:ood} positions. A human facing a randomized board would therefore plausibly require more time and might select weaker moves. We argue that this objection applies with very different force to our two metrics, and we scope our claims accordingly.

\emph{Move quality.} Finding a strong move in an unfamiliar position plausibly requires more deliberation from any agent. Without a human baseline on our specific positions, we cannot rule out that time-constrained humans would also show some degradation in move quality on \gls{acr:ood} positions. We therefore regard the \gls{acr:acpl} results as \emph{consistent with}, but not conclusively isolated from, a generic difficulty-under-novelty effect.

\emph{Rule adherence.} Legality differs categorically from move quality. The rules that determine whether a move is legal are fixed and identical across all positions. Applying them requires reconstructing the board state, but this procedure is fully specified, bounded, and requires no strategic judgment or pattern knowledge. An agent that has internalized the rules should therefore become slower, but not less rule-compliant, on unfamiliar positions. Evidence from a simplified setting is consistent with this expectation for humans: \citet{reingold2023chess} tested expert and novice human players on a double-check detection task, rendered either in the familiar symbol notation or in an unfamiliar letter notation. Error rates remained below $1\%$ for both groups in both notations. The unfamiliar notation slowed responses but did not reduce accuracy, indicating that human players retain their ability to adhere to a fixed rule set, even under aggravated circumstances.

By contrast, the \glspl{acr:llm} in our study propose illegal moves at sharply increasing rates as positions become less familiar, and additional reasoning does not eliminate this failure: GPT-5 with moderate reasoning still proposes illegal moves in more than $10\%$ of \gls{acr:ood} positions despite using roughly five times as many reasoning tokens as in \gls{acr:wd} positions. Because verifying move legality is a bounded procedure, additional deliberation should suffice to resolve it, as additional time would for a human player. We therefore regard the illegal-move results as the more robust evidence that current \glspl{acr:llm} fail to apply the rules of chess systematically outside familiar configurations.

A dedicated human study on our position set would provide a direct test of both claims and constitutes a valuable avenue for future work.

\section{Extended related work}\label{sec:extended_related_work}

This appendix contrasts our approach with closely related work at the intersection of language models and chess.

\textbf{\citet{ruoss2024amortizedplanninglargescaletransformers}} introduce ChessBench, a large-scale dataset for chess, and train specialized transformers using supervised learning on Stockfish evaluations. While they do not reach superhuman performance, they demonstrate that playing strength scales with model and dataset size. Our work differs from theirs in several critical ways. \citet{ruoss2024amortizedplanninglargescaletransformers} train a specialized transformer via supervised learning on Stockfish-generated labels to maximize overall chess performance. In contrast, we evaluate general-purpose frontier \glspl{acr:llm} zero-shot to disentangle memorization from reasoning. Further, their evaluation relies on data derived from human play, meaning the models are tested on naturally occurring board states. In this regime, pattern memorization remains an extremely effective heuristic. By limiting their evaluation to the region of the chess state space frequently encountered in human play, their setup differs from ours. Given their findings, it would be a highly interesting avenue for future research to evaluate their specialized models on structurally novel \gls{acr:ood} positions to observe if extensive chess-centric training helps mitigates the generalization gap we observed in frontier \glspl{acr:llm}.

\textbf{\citet{wang2024_finetuning-for-chess}} introduce MATE, a dataset of approximately one million positions from chess games and puzzles in which candidate moves are annotated with explanations of long-term strategy and short-term tactics. They fine-tune LLaMA-3-8B on MATE and show that the fine-tuned model outperforms commercial \glspl{acr:llm}, including GPT, Claude, and Gemini models, and that language explanations improve this ability. Their setting differs from ours in that models are fine-tuned on chess data, the task is a choice between provided candidate moves (which removes the need to generate legal moves), and positions stem from human play and puzzles. We instead evaluate unconstrained, zero-shot move generation across conditions that differ in the density of relevant priors.

\textbf{\citet{hwang2025_knowledge-dist}} investigate whether \glspl{acr:llm} can develop strategic reasoning through reinforcement learning on chess, using a chess-pretrained action-value network to provide dense rewards on move quality, a form of knowledge distillation. All models plateau far below expert level, which the authors attribute to a deficit in the pretrained models' internal understanding of chess that reinforcement learning alone may not overcome. This conclusion is complementary to ours: we characterize the zero-shot competence of frontier models directly and find that it is concentrated in regions of the state space that are densely covered by likely training data.

\textbf{\citet{chess-llm-performance}} study ChatGPT's chess abilities using metrics for both the legality and the quality of its moves, and report limitations in its comprehension of the formal board representation, identifying rule adherence as a central weakness. We extend this line of work by showing that rule adherence degrades systematically as positions become less familiar, and that this pattern persists across three GPT generations, two further model lineages, and reasoning-augmented inference.

\textbf{\citet{kolasani2025llmchessbenchmarkingreasoning}} introduce LLM CHESS, an agentic benchmark in which more than 50 models play complete games against a random opponent and can retrieve the board state and the list of legal moves through tool calls. Elo estimates are derived from games against a chess engine for a subset of models. The authors report that state-of-the-art models struggle to complete games or to win consistently, and reasoning models separate clearly from non-reasoning models. In contrast to our work, their benchmark measures end-to-end agentic play rather than performance as a function of positional familiarity.

\clearpage
\section{Additional analyses}\label{sec:additional_analyses}

\subsection{Centipawn loss distributions}\label{sec:cpl_dist}

\begin{figure}[h!]
    \centering
    \includegraphics[width=1\linewidth]{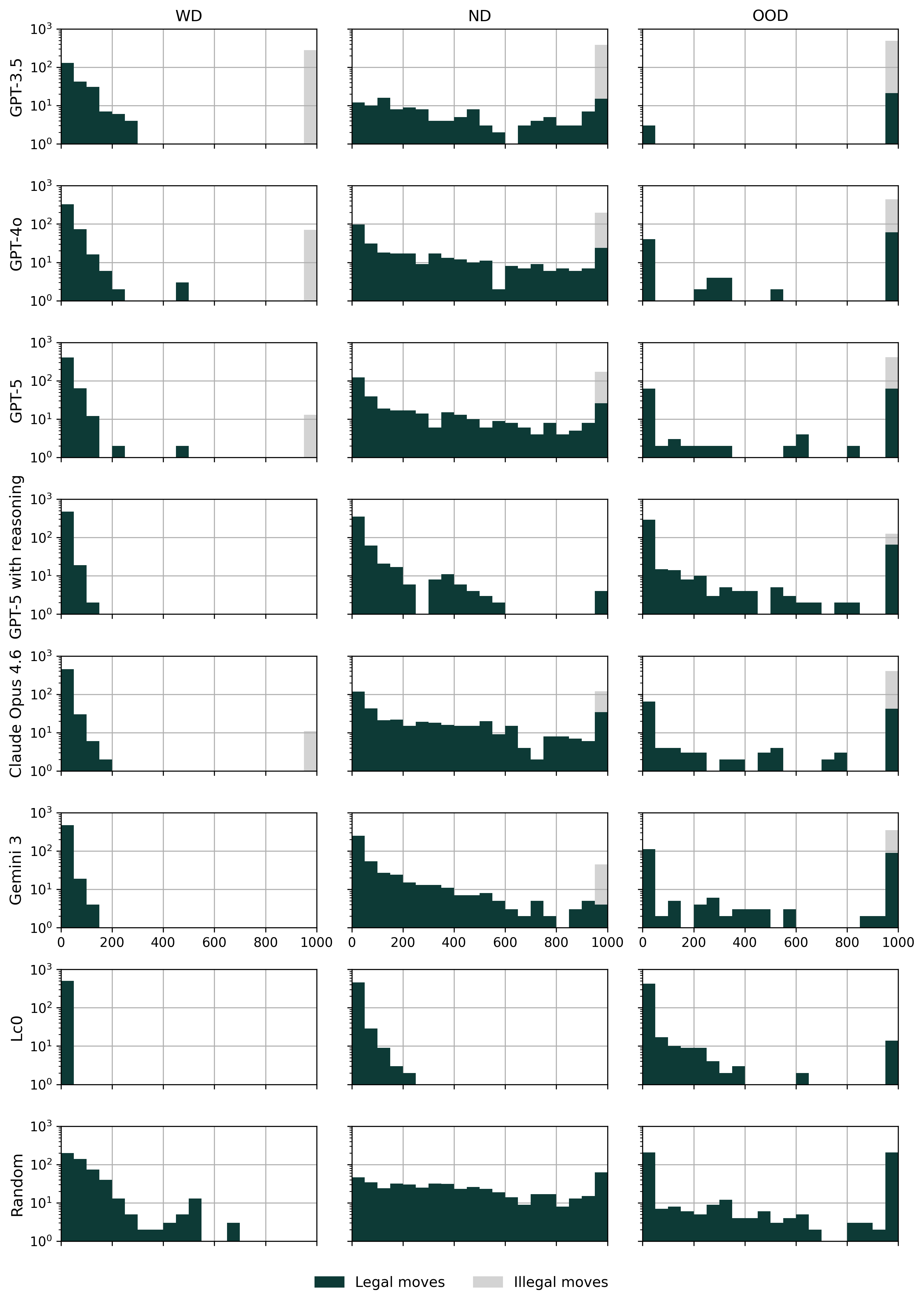}
    \caption{Centipawn loss distributions by model and condition. WD = within-distribution, ND = near-distribution, OOD = out-of-distribution.}
    \label{fig:cpl_distributions}
\end{figure}

\subsection{Split diagnostics}\label{sec:split_diagnostics}

Table~\ref{tab:split_diagnostics} characterizes the three conditions. \gls{acr:wd} and \gls{acr:nd} positions retain near-complete and balanced material, whereas \gls{acr:ood} positions contain substantially less material and exhibit a much larger variance in the number of legal moves.

\begin{table}[h]
  \centering
  \small
  \caption{Split diagnostics (mean $\pm$ SD over 500 positions per condition). Material in pawn units (P\,=\,1, N\,=\,B\,=\,3, R\,=\,5, Q\,=\,9; kings excluded). Legal moves are counted for the side to move. In check: share (count) of positions in which the side to move is in check.}
  \label{tab:split_diagnostics}
  \begin{tabular*}{\textwidth}{@{\extracolsep{\fill}}lcccc@{}}
    \toprule
    Condition & Legal moves & White material & Black material & In check \\
    \midrule
    WD  & $26.27 \pm 4.42$  & $38.98 \pm 0.13$ & $38.99 \pm 0.11$ & $0.4\%$ (2)   \\
    ND  & $31.21 \pm 7.68$  & $37.76 \pm 2.20$ & $37.84 \pm 2.30$ & $3.4\%$ (17)  \\
    OOD & $19.75 \pm 18.85$ & $24.36 \pm 4.69$ & $24.99 \pm 4.74$ & $54.0\%$ (270) \\
    \bottomrule
  \end{tabular*}
\end{table}

Because OOD positions are generated by random piece placement, constrained only so that the side not to move is not in check, the side to move is in check in 54.0\% of OOD positions, compared with 0.4\% of WD and 3.4\% of ND positions (Table~\ref{tab:split_diagnostics}). Being in check is strongly associated with illegal moves. In OOD positions in check, illegal-move rates range from 21.9\% (GPT-5 with moderate reasoning) to 99.3\% (GPT-3.5), compared with 0.9\% to 87.8\% in OOD positions not in check, so in-check positions account for 57--97\% of all illegal moves on OOD positions. The same pattern holds in the 17 ND positions in check (e.g., 64.7\% vs.\ 27.5\% for GPT-5 with minimal reasoning)\footnote{The sole exception is GPT-5 with moderate reasoning, as it did not play any illegal moves in the ND condition.}. Much of the illegal-move gradient on OOD positions thus reflects the models' inability to recognize that they are in check, or to respond to it with a legal move. The gradient persists without check, however: among positions not in check, illegal-move rates still rise from ND to OOD for all models (e.g., from 32.7\% to 51.7\% for GPT-4o), although by a smaller margin.

\subsection{Robustness to the output notation}\label{sec:uci_ablation}

Each model was tasked to return its recommendation in both SAN and UCI notation within the same response (Appendix~\ref{sec:prompt}). Table~\ref{tab:uci_ablation} reports results when moves are taken from the UCI field instead of the SAN field. Moves were parsed as in the main analysis (Section~\ref{sec:methodology}). All qualitative findings replicate: For every model and reasoning setting, the rate of legal moves is lowest and the \gls{acr:acpl} highest in \gls{acr:ood} positions, both including and excluding illegal moves. Successive GPT generations generally improve, with diminishing gains in the \gls{acr:ood} condition. Reasoning improves performance in all conditions. GPT-5 with moderate reasoning still proposes illegal moves in $12.4\%$ of \gls{acr:ood} positions.

\begin{table}[h]
  \centering
  \small
  \caption{Results for moves taken from the UCI field, using Stockfish~17.1 at search depth~20. ACPL including illegal moves assigns a loss of $1{,}000$ to illegal moves; ACPL excluding illegal moves is computed over legal moves only.}
  \label{tab:uci_ablation}
  \begin{tabular*}{\textwidth}{@{\extracolsep{\fill}}llrrr@{}}
    \toprule
    Model & Condition & Legal (\%) & ACPL (incl.\ illegal) & ACPL (excl.\ illegal) \\
    \midrule
    GPT-3.5          & WD  & 37.2  & 648.17 & 54.22  \\
                     & ND  & 17.6  & 893.35 & 394.05 \\
                     & OOD & 4.2   & 989.46 & 749.05 \\
    \midrule
    GPT-4o           & WD  & 85.6  & 174.52 & 35.65  \\
                     & ND  & 72.0  & 514.16 & 325.22 \\
                     & OOD & 32.4  & 849.37 & 535.08 \\
    \midrule
    GPT-5 (minimal)  & WD  & 97.0  & 55.53  & 26.32  \\
                     & ND  & 79.0  & 440.76 & 292.10 \\
                     & OOD & 35.0  & 838.03 & 537.23 \\
    \midrule
    GPT-5 (moderate) & WD  & 100.0 & 9.29   & 9.29   \\
                     & ND  & 100.0 & 69.54  & 69.54  \\
                     & OOD & 87.6  & 286.80 & 185.84 \\
    \bottomrule
  \end{tabular*}
\end{table}

\section{Societal impact}\label{sec:societal_impact}

This research highlights a generalization gap in \glspl{acr:llm} within the domain of chess, suggesting that their impressive performance may stem more from extensive pattern recall than from adaptive, first-principles reasoning. By demonstrating that model reliability can decline when encountering novel configurations, these findings encourage a cautious approach to deploying LLMs in specialized or rapidly evolving fields where historical data may not be a perfect guide.

\section{Hardware specification}\label{sec:hardware}
All experiments were performed on a 2024 MacBook Air with an Apple M3 processor.



\end{document}